\documentclass[10pt,twocolumn,letterpaper]{article}

 \usepackage{cvpr}              
\definecolor{cvprblue}{rgb}{0.21,0.49,0.74}
\usepackage[pagebackref,breaklinks,colorlinks,allcolors=cvprblue]{hyperref}

\usepackage{graphicx}
\usepackage{booktabs}

\usepackage{wrapfig}
\usepackage{algorithm}
\usepackage{algorithmic}
\usepackage{amssymb} 

\usepackage{booktabs}
\usepackage[table]{xcolor}  
\definecolor{RowColor}{HTML}{E7F3F7}
\definecolor{HeaderColor}{HTML}{E7E6E6}
\usepackage{caption}

\makeatletter
\g@addto@macro\@maketitle{%
  \vspace*{-0.4\baselineskip}
  \begingroup
  \centering
  \includegraphics[width=\linewidth]{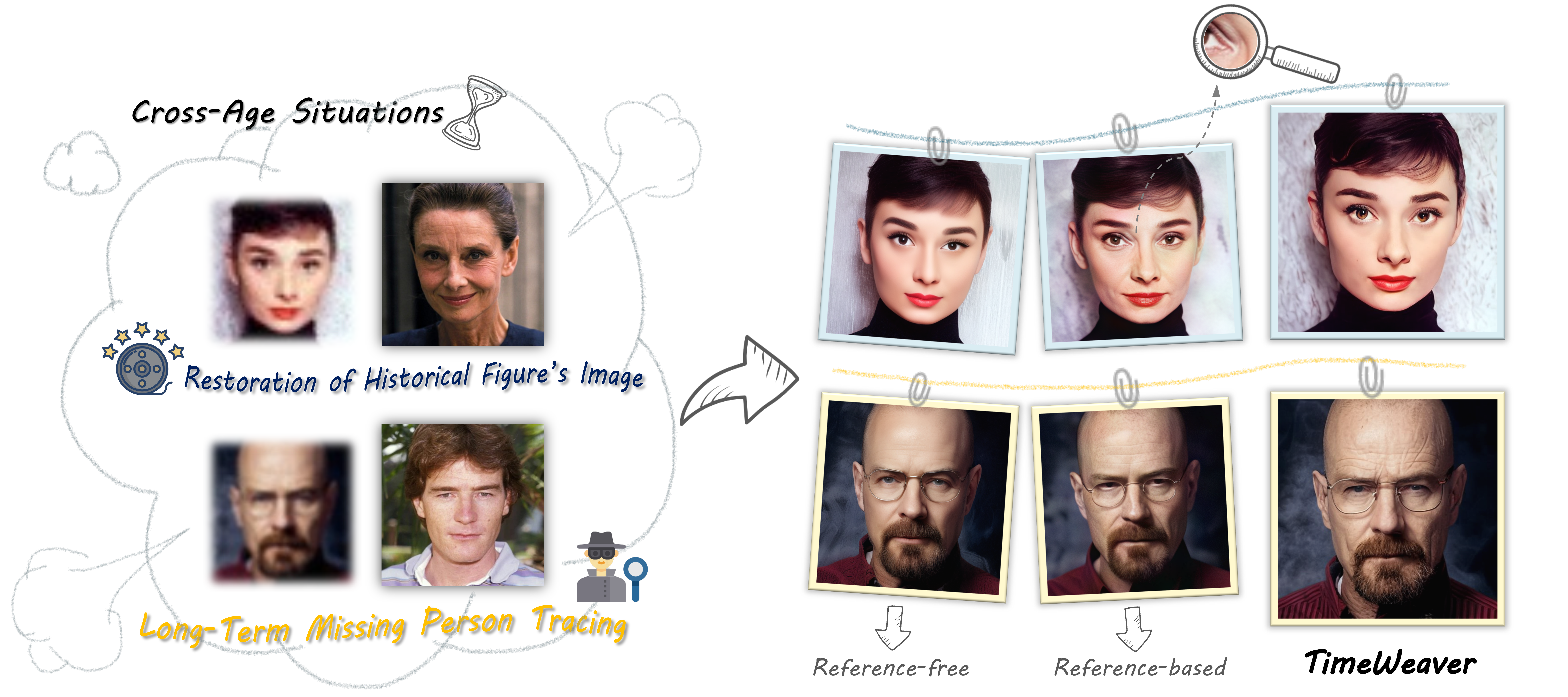}%
  \vspace*{-0.4\baselineskip}
  \captionof{figure}{Given degraded inputs and reference images with large age gaps, reference-free methods \cite{diffbir} struggle to preserve identity, while reference-based methods \cite{faceme} fail to main age fidelity. \textbf{TimeWeaver} achieves identity-faithful and age-consistent restoration.}
  \label{fig1}
  \vspace*{1\baselineskip}%
  \par
  \endgroup
}
\makeatother

\def\paperID{16080} 
\def\confName{CVPR}
\def\confYear{2026}

\title{TimeWeaver: Age-Consistent Reference-Based Face Restoration\\ with Identity Preservation}

\author{Teer Song$^{1}$,
Yue Zhang$^{1}$,
Yu Tian$^{2}$,
Ziyang Wang$^{1}$,
Xianlin Zhang$^{1}$,
Guixuan Zhang$^{1}$,\\
Xuan Liu$^{3}$,
Xueming Li$^{1}$,
Yasen Zhang$^{4}$ \\
$^{1}$Beijing University of Posts and Telecommunications 
\quad $^{2}$Tsinghua University \\
$^{3}$Minzu University of China \quad
  $^{4}$Xiaomi Corporation \\
}

\begin{document}

\maketitle
\begin{abstract}
Recent progress in face restoration has shifted from visual fidelity to identity fidelity, driving a transition from reference-free to reference-based paradigms that condition restoration on reference images of the same person. However, these methods assume the reference and degraded input are age-aligned. When only cross-age references are available, as in historical restoration or missing-person retrieval, they fail to maintain age fidelity. To address this limitation, we propose TimeWeaver, the first reference-based face restoration framework supporting cross-age references. Given arbitrary reference images and a target-age prompt, TimeWeaver produces restorations with both identity fidelity and age consistency. Specifically, we decouple identity and age conditioning across training and inference. During training, the model learns an age-robust identity representation by fusing a global identity embedding with age-suppressed facial tokens via a transformer-based ID-Fusion module. During inference, two training-free techniques—Age-Aware Gradient Guidance and Token-Targeted Attention Boost—steer sampling toward desired age semantics, enabling precise adherence to the target-age prompt. Extensive experiments show that TimeWeaver surpasses existing methods in visual quality, identity preservation, and age consistency.
\end{abstract} 
\section{Introduction}
\label{sec:intro}
Fidelity lies at the core of image restoration, and this demand becomes even more stringent when it comes to faces. Beyond reconstructing global facial structure or enhancing visual quality, face restoration places growing emphasis on identity fidelity—that is, whether the restored face still looks like the same person. Recent research has therefore shifted from reference-free to reference-based paradigms. Compared to reference-free methods \cite{codeformer,GFP-GAN, difface, dr2, diffbir} that restore faces directly from degraded inputs, reference-based methods \cite{DMDNet, ref-ldm, pfstorer,faceme,instantrestore} leverage high-quality reference images of the same person to provide identity cues and have become the prevailing approach. 

However, this paradigm suffers from a blind spot. In many practical situations, the available reference images are not contemporaneous with the target face to be restored. For instance, in historical image restoration, a deteriorated photograph may need to be enhanced for archival preservation, while the only available reference is a portrait of the same individual taken decades apart. Similar situations arise in tracing long-term missing persons or in forensic investigation. In such cross-age scenarios, one can’t help but wonder—can the current reference-based methods still deliver faithful restoration?

Unfortunately, as shown in Fig.~\ref{fig1}, although the restored image may maintain identity fidelity, it appears to directly copy age-related features (\emph{e.g.}, skin texture, wrinkles, facial fullness) from the reference. If the restored face resembles the person from the reference period rather than the time of capture, how credible is the result for identification or historical interpretation? Thus, it can be asserted that such restoration fails to maintain age fidelity. 

To tackle this pain point, we introduce TimeWeaver, to the best of our knowledge the first reference-based face restoration framework capable of handling cross-age references and producing identity- and age-faithful restorations. We begin with a reference-free restoration model \cite{diffbir} built upon Stable Diffusion \cite{sd} and extend it by injecting identity features from reference images and incorporating a user-specified target age in form of text prompt as condition. Our study centers on two key questions: \emph{(1) how to extract reliable identity features from the reference while mitigating the influence of its age traits, and (2) how to generate precise target-age semantics.} Jointly addressing the two questions—namely, training the base model with both identity and age conditions in a supervised manner—is problematic. Identity and age are inherently entangled, implicit age cues in reference images may conflict with the target age and misguide the model’s learning objective, while the scarcity of cross-age identity-paired training data further limits the model’s ability to learn disentangled representations (see a detailed dataset analysis in the Appendix B).Therefore, our key solution is to decouple the processing of identity and age conditions across training and inference stages. 

During training, we focus solely on identity preservation, addressing the first question. Inspired by personalized generation approaches \cite{instantid, face_adapter, pulid}, we adopt a pretrained face recognition model \cite{arcface} to extract a global age-robust identity embedding, and complement it with detailed semantics derived as age-suppressed facial tokens from the CLIP-ViT encoder \cite{clip}. Specifically, we design a feature-extraction scheme that reduces the contribution of age-sensitive regions in the CLIP branch, encouraging the resulting tokens to carry identity-relevant facial structures rather than age-specific attributes such as skin texture or wrinkle patterns. In addition, a transformer-based \cite{transformer} ID-Fusion module fuses the two feature sets into a compact set of age-irrelevant identity tokens, which serve as image prompt and are injected into the base model via decoupled cross-attention \cite{ip}, with a unified prompt ``photo of a person'' to guide the restoration.

During inference, we tackle the second question by introducing age semantics as a text-driven editing signal. We describe this as editing because, after training with identity conditioning, the model’s responsiveness to age prompt attenuates and sometimes fails to follow the specified age (as shown in Fig.~\ref{fig3}). To address this, we propose two training-free techniques that jointly enable reliable age control during generation: Age-Aware Gradient Guidance (AAGG) computes an age-directional gradient in the latent space using paired prompts (\emph{e.g.}, “photo of a person” vs. “photo of a 24-year-old person”) and leverages it to refine the latent through an optimization procedure toward the desired age manifold. In parallel, Token-Targeted Attention Boost (TTAB) uses the attention map of age-specific tokens as a modulation weight to concentrate updates on regions that express age semantics. Together, they achieve global semantic steering with spatial selectivity, enabling precise adherence to target-age prompt without additional training.

In practice, TimeWeaver accepts any number of reference images without age-span constraints, along with a user-specified target-age text corresponding to the degraded, producing restorations with both identity fidelity and age consistency. Extensive experiments show that TimeWeaver establishes a new state-of-the-art, achieving superior visual quality, identity similarity, and age consistency. Our key contributions are as follows:
\begin{itemize}
    \item We present the first reference-based face restoration framework capable of handling cross-age references without age-span constraints.
    \item We propose a disentangled training–inference strategy that learns identity conditioning during training and enforces age semantics at inference, effectively mitigating identity–age conflicts and cross-age data scarcity.
    \item We develop an age-robust identity representation by fusing global identity features and semantic facial details via the proposed ID-Fusion, and introduce two training-free techniques to achieve precise target-age guidance.
    \item Experiments on both same-age and cross-age benchmarks show that our method outperforms existing approaches in visual quality, identity similarity, and age consistency.
\end{itemize}

\section{Related Work}
\label{sec:rel}

\subsection{Reference-free Face Restoration}
 Reference-free face restoration refers to conventional blind face restoration approaches that aim to recover high-quality face images from the degraded input under unknown degradation, without reference images provided. Most GAN-based methods \cite{codeformer, vqfr, GFP-GAN, copyornot} exploit pre-trained models such as StyleGAN2 \cite{stylegan2} or a learned VQ codebook of facial features as priors to directly synthesize realistic facial details. Recently diffusion-based methods \cite{dr2, difface, pgdiff, diffbir} exploit the strong generative priors of diffusion models for high-fidelity restoration. However, since these methods proceed without reference images, the restored results may deviate from the authentic identity, particularly under severe degradation.

\subsection{Reference-based Face Restoration}
Reference-based face restoration aims to enhance identity fidelity by leveraging high-quality reference images of the same identity. Alignment-based methods \cite{ASFFNet, DMDNet} rely on enforcing geometric alignment between the reference and degraded input to transfer facial details. Diffusion-based method \cite{pfstorer} learns personalized representations from a few reference samples but requires per-identity tuning. To avoid test-time tuning, recent approaches \cite{faceme, restorerid} train an identity encoder to extract identity embeddings and use them as conditioning signals, achieving tuning-free restoration. However, these methods are effective only when the reference and degraded input have a minimal age gap and fail to maintain age fidelity in cross-age scenarios.

\subsection{Image Editing with Diffusion Models}
The strong text priors of diffusion models have propelled a surge of prompt-to-prompt editing methods \cite{p2p, dds, null}. Building on this trend, Score Distillation Sampling \cite{sds} first demonstrated how 2D diffusion outputs can serve as gradient guidance to drive 3D scene generation, and subsequent variants such as \cite{dds, cds, dreamsteerer} extended this idea to image editing field, enabling training-free and inversion-free local edits. In personalized face image generation, methods like \cite{masterweaver, face2diffusion} further apply this strategy by leveraging diffusion priors to define editing-direction losses that enhance fine-grained facial attribute control capability. Inspired by these advances, we extend such editing strategies to the task of age-aligned face restoration, allowing precise age control while preserving identity.
\section{Method}

\subsection{Overview}
\begin{figure*}[t]
  \centering
  \includegraphics[width=0.99\linewidth]{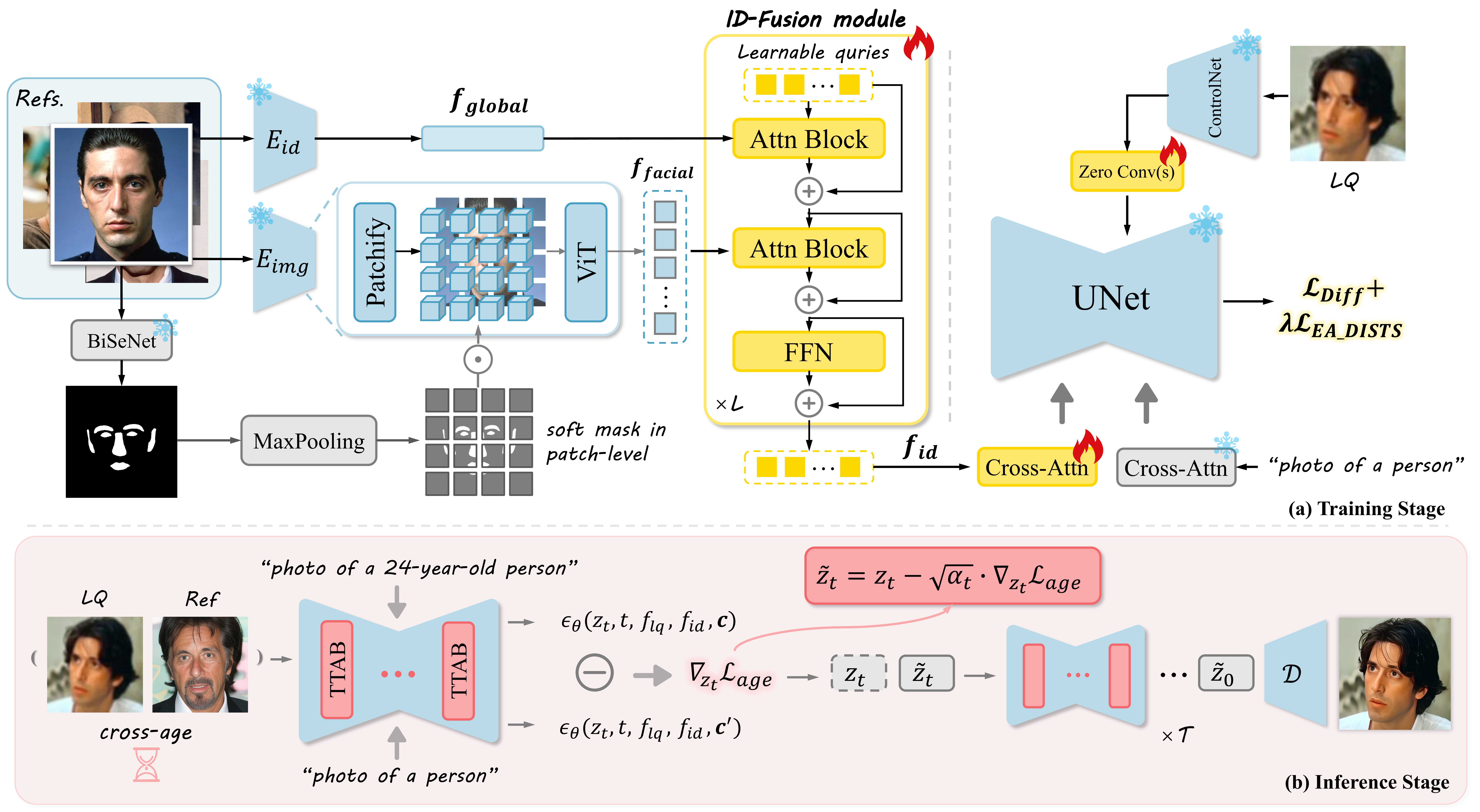} 
  \vspace{-0.3em}
  \caption{\textbf{Framework of TimeWeaver.} (a) During training, we extract a global identity embedding using ArcFace \cite{arcface} and age-suppressed facial tokens using CLIP-ViT \cite{clip}. The ID-Fusion fuses them and injects the output into UNet as the identity condition. (b) During inference, the framework computes an age-aware gradient with and without the age condition to refines the latent along the denoising process, assisted by TTAB technique to achieve age-controllable restoration.  }
  \vspace{-1em}
  \label{fig2}
\end{figure*}

In this section, we provide an overview of TimeWeaver. Our method builds upon DiffBIR \cite{diffbir}, a reference-free restoration model that injects the degraded image into the diffusion process via ControlNet \cite{controlnet} to follow the original structure. On top of this backbone,  we extend it to a higher-level objective—identity-preserving and age-consistent restoration—by introducing reference images and a target-age text as conditions. The workflow unfolds in two stages. We start by focusing on training a reference-based restoration model conditioned on age-robust identity features extracted from reference images. Once the model has learned \emph{who} the person is, we shift our focus in the inference stage to \emph{when}, enabling age control to guide the restoration toward the desired age. Details are discussed in the following subsections, and the framework is shown in Fig.~\ref{fig2}.

\subsection{Training Stage: Identity Preservation}
Due to the lack of sufficient cross-age identity datasets and entangled feature conflicts, achieving identity-faithful and age-aligned restoration in a fully supervised manner is challenging (see Appendix B for datasets analysis). To address this, we adopt a deliberate strategy by first training a reference-based face restoration model on available identity datasets \cite{vggface2, DMDNet}, establishing a solid foundation for identity fidelity. In the remainder of this section, we describe how identity features are extracted, fused, and injected, followed by the training scheme.

\textbf{Age-Robust Identity Embedding.} Inspired by personalized image generation methods \cite{instantid, facestudio, pulid, portraitbooth}, we use a pretrained face recognition (FR) model \cite{arcface} denoted $E_{id}(\cdot)$ to extract a global identity representation. Trained with an identity-discriminative objective, the FR model is designed to produce consistent embeddings for the same person even when age varies, yielding identity features naturally decoupled from age \cite{arcface, mtlface, elasticface}. Given reference images $\{I_i\}_{i=1}^N$, we obtain each identity embedding $e_i=E_{id}(I_i)\in \mathbb{R}^{512}$, and compute their mean to form a global age-robust identity embedding $
f_{global} = \frac{1}{N} \sum_{i=1}^{N} e_i\in \mathbb{R}^{512}
$. 

\textbf{Age-Suppressed Facial Tokens.} Although the FR model provides a stable global identity representation, recent studies (e.g., FaceID-Plus \cite{ip}) suggest that CLIP features can further enrich facial semantics and perceptual details. We therefore leverage a pretrained CLIP-ViT encoder \cite{clip} to extract fine-grained facial cues. However, as a general-purpose visual encoder, CLIP may encode salient age-related cues such as skin texture and wrinkles. To suppress such age leakage, we introduce a masking-based token reweighting mechanism that biases the representation toward identity-bearing facial organ structures, which remain relatively stable across ages, while attenuating skin-dominant regions where age-related changes are most prominent.

Specifically, we use a face parser \cite{bisenet} to segment organ regions (eyes, nose, mouth, eyebrows, and ears), and merge them into a binary mask $M\in\{0,1\}^{H\times W}$. If parsing fails due to severe occlusion or extreme pose, we discard the corresponding reference image for this instance. Instead of masking the image at the pixel-level—which would shift the input distribution away from CLIP's training data—we apply masking at the patch-level (\emph{i.e.}, after patch projection and before the ViT encoder). At this point, the reference image is patchified into tokens $x \in \mathbb{R}^{(p \times p)\times d_v}$ with position encoding, where $p \times p$ is the number of patches and $d_v$ is the CLIP visual embedding dimension.  The mask $M$ undergoes the same geometric preprocessing and is converted to patch-level soft values via non-overlapping max pooling with kernel $p$, followed by a linear gain mapping and unit-mean normalization:
\begin{equation}
m = \mathrm{MaxPool}_{p}(M), \qquad m \in [0,1]^{p \times p}
\end{equation}
\vspace{-0.5em}
\begin{equation}
w_i = \frac{1 + \beta m_i}{\frac{1}{p^2}\sum_{j=1}^{p^2}(1+\beta m_j)}, \qquad w \in \mathbb{R}^{p \times p},
\end{equation}
where $\beta$ governs the enhancement strength, ensuring $\frac{1}{1 + \beta} \;\le\; w_i \;\le\; 1 + \beta$. Applying $w$ to the patch tokens gives $x'=x\odot w$, so facial-organ patches (with larger $m_i$) are relatively amplified, while non-ROI regions are attenuated. Feeding $x'$ into the ViT encoder, we extract penultimate-layer features \cite{consistentid, uniportrait} and average over the effective reference set to obtain the age-suppressed facial tokens: $f_{facial}=\frac{1}{N} \sum_{i=1}^{N} E_{img}(x')\in \mathbb{R}^{(p \times p) \times d_v}$, where $N$ is the number of valid reference images for the instance. 

\textbf{ID-Fusion.} To integrate the global identity embedding $f_{global}$ and age-suppressed facial tokens $f_{facial}$, we introduce a lightweight module named ID-Fusion. Our ID-Fusion introduces $n$ learnable query tokens and stacks $L$ layers. Each layer comprises two cross-attention blocks followed by an FFN, enabling the queries to attend to $f_{global}$  and $f_{facial}$ (as key/value) to integrate identity consistency and fine-grained details. To improve robustness against mask extraction errors, we randomly drop the second cross-attention block (corresponding to $f_{facial}$) with probability 15\%.  After $L$ layers, the refined queries are projected to the UNet cross-attention dimension $d_c$, yielding $f_{id} \in \mathbb R^{n \times d_c} $, which are then injected via decoupled cross-attention \cite{ip}.

\textbf{Training Scheme.}  We adopt the pretrained ControlNet \cite{controlnet} from DiffBIR \cite{diffbir} to encode the degraded image as $f_{lq}$. To accommodate identity branch, we unfreeze the zero-convolution layers and jointly train with ID-Fusion and newly added cross-attention layers. We use a unified text prompt $ \text{``photo of a person''}$ encoded as $f_t$, and the diffusion loss is:
\begin{equation}
\mathcal{L}_{\text{Diff}} = \mathbb{E}_{z_t,\, t,\,f_{lq},\, f_{{id}},\, f_{{t}},\, \epsilon} 
 \left\| \epsilon - \epsilon_\theta(z_t,\,t,\,f_{lq},\, f_{id},\,f_t) \right\|_2^2 ,  
\end{equation}
where ${z}_t$ denotes the latent at timestep $t$, $\epsilon_\theta(\cdot)$ the noise predictor and $\epsilon \sim \mathcal{N}(0, I)$. To mitigate the over-smoothing outputs of diffusion-based restoration, especially in hair and skin regions, we follow OSDFace \cite{osdface} and adopt Edge-Aware DISTS (EA-DISTS) \cite{ea_dist} as a perceptual loss:
\begin{equation}
\mathcal{L}_{\text{EA-DISTS}}(\hat{I}, I_{gt})
= \mathcal{L}_{\text{DISTS}}(\hat{I}, I_{gt})
+ \mathcal{L}_{\text{DISTS}}\bigl( \mathcal{S}(\hat{I}), \mathcal{S}(I_{gt}) \bigr),
\end{equation}
where $I_{gt}$ is the ground truth image, $\hat{I}$ the decoded prediction, and $\mathcal{S}(\cdot)$ the Sobel operator. The total loss is given by: 
\begin{equation}
\mathcal{L}_{total} = \mathcal{L}_{\text{Diff}}+\lambda\mathcal{L}_{\text{EA-DISTS}},
\end{equation}
where $\lambda$ is a scale factor. Under a warm-up strategy, EA-DISTS is activated only in the later training phase to avoid early-stage instability. 

\subsection{Inference Stage: Age Control Generation}
\begin{figure} 
  \centering
  \includegraphics[width=\linewidth]{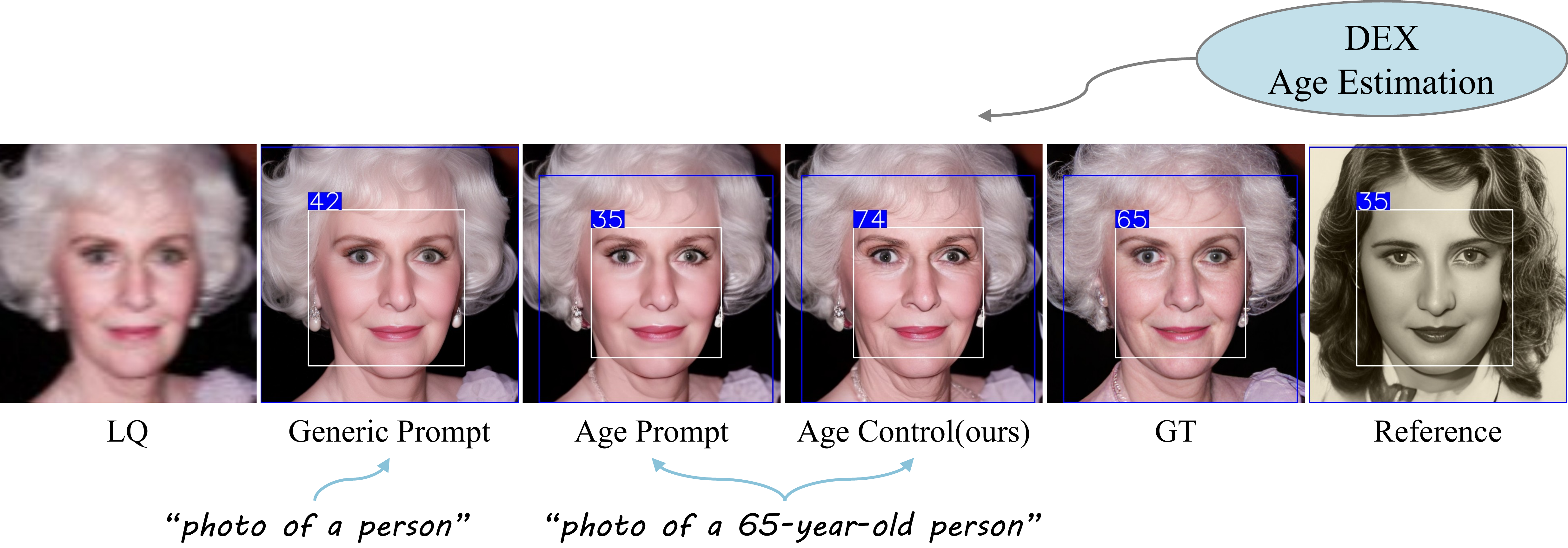}
  \vspace{-1.5em}
  \caption{Visual comparison of different inference strategies with
DEX \cite{dex} age estimates. }
  \label{fig3}
  \vspace{-0.5em}
\end{figure}
After obtaining a restoration model with strong identity preservation, one might naturally attempt to leverage age prompt to query the model for age-specific generation. However, empirical observations shows that strengthening the image prompt compromises text controllability, as illustrated in Fig.~\ref{fig3}, this trade-off also noted in related works \cite{attndreambooth, pulid}. We therefore introduce two training-free techniques that work together to reactivate the model’s semantic responsiveness, enabling precise and distinct age control during inference. The whole algorithm is shown in Algorithm.~\ref{alg1}.

\textbf{Age-Aware Gradient Guidance.} We first introduce an age-specific prompt of the form ``photo of a [$\tau$]-year-old person'', where $\tau$ is the input target age in numerals. This follows prior work \cite{fading} showing that numeral-based expressions better capture age characteristics than coarse prompts (\emph{e.g.}, ``man in his thirties'') or vague descriptors. Then following the score-based view of diffusion models \cite{score2, score}, the UNet can be viewed as a conditional score estimator whose output approximates the gradient of the log-density:
\begin{equation}
\nabla_{z_t}\log p_t\big(z_t \mid f_{lq}, f_{id}, c\big)
\propto -\,
\epsilon_\theta\big(z_t, f_{lq}, f_{id}, c, t\big),
\end{equation}
where $c$ is text prompt. Motivated by this view, we isolate the age attribute by taking the score difference under age-specific prompt $c'$ and generic prompt $c$ (``photo of a person'') to provide an age-aware gradient defined as:
\begin{equation}
\nabla_{z_t} \mathcal{L}_{\text{age}} = \epsilon_\theta(z_t,\,f_{lq},\, f_{id},\,c', \, t) - \epsilon_\theta(z_t,\,f_{lq},\, f_{id},\,c, \, t).
\end{equation}
This residual captures the direction pointing from the model’s prediction on $z_t$ conditioned on $c'$ to the prediction conditioned on $c$, canceling out components unrelated to the age attribute---such as identity features---thus enabling high-level conceptual guidance focused purely on age semantics. Then, we leverage it to refine the current latent  $z_{t}$, obtaining the updated $\tilde{z}_{t}$:
\begin{equation}
\tilde{z}_{t} = z_{t} - \sqrt{\alpha}_t \cdot \nabla_{z_t} \mathcal{L}_{\text{age}},
\end{equation}
the modulation term $\sqrt{\alpha_t}$  is consistent with its definition in typical diffusion process \cite{ddpm} and serves to adaptively regulate the strength of the guidance across different timesteps. The updated $\tilde{z}_{t}$ is then used to compute ${z}_{t-1}$ along the DDIM sampling trajectory \cite{ddim} until reaching ${z}_{0}$. By incrementally nudging $z_{t-1}$ along the age-specific semantic direction, the generated image better aligns with the target age while preserving identity consistency.

\begin{algorithm}[t]
\caption{Age Control Generation at Inference.}
\label{alg1}
\textbf{Input:} image conditions $f_{lq}$, $f_{id'}$, source prompt $f_t$, target-age prompt $f_t'$, optimization steps $N$, step size $\eta$ \\
\textbf{Output:} restored image $\hat I$
\begin{algorithmic}[1]
\STATE Sample $z_T \sim \mathcal{N}(0, I)$
\FOR{$t = T$ \textbf{to} $1$}
  \FOR{$n = 1$ \textbf{to} $N$}
    \STATE $z_t \leftarrow z_t.\texttt{detach}().\texttt{requires\_grad}()$

    \STATE {\small\textit{\# UNet forward (generic prompt; TTAB off)}}
    \STATE $\epsilon_{\text{src}} \leftarrow \epsilon_\theta(z_t,\, t,\, f_{lq},\, f_{id'},\, f_t;\, \text{no TTAB})$

    \STATE {\small\textit{\# UNet forward (age prompt; TTAB on)}}
    \STATE $\epsilon_{\text{trg}} \leftarrow \epsilon_\theta(z_t,\, t,\, f_{lq},\, f_{id'},\, f_t';\, \text{TTAB})$

    \STATE $\Delta\epsilon \leftarrow (\epsilon_{\text{trg}} - \epsilon_{\text{src}}).\texttt{detach}()$
    \STATE $\mathcal{L}_{\text{age}} \leftarrow (\Delta\epsilon \cdot z_t).\mathrm{mean}()$
    \STATE $z_{t} \leftarrow z_{t} - \eta \cdot \sqrt{\alpha_t} \cdot \nabla_{z_t}\mathcal{L}_{\text{age}}$
  \ENDFOR

  \STATE $z_{t-1} \leftarrow 
      \sqrt{\alpha_{t-1}}
      \left(
        \frac{z_t - \sqrt{1-\alpha_t}\,\epsilon_\theta(z_t,\, t,\, f_{lq},\, f_{id'},\, f_t';\, \text{TTAB})}
        {\sqrt{\alpha_t}}
      \right)$
\ENDFOR
\STATE \textbf{return} $\hat I \leftarrow \mathcal{D}(z_0)$
\end{algorithmic}
\end{algorithm}

\textbf{Token-Targeted Attention Boost.} The latent updates in AAGG provides global guidance, which inevitably spreads gradient over the whole image and may induce texture fluctuations in non-age-related regions such as the background. 
\begin{figure}
  \centering
  \includegraphics[width=\linewidth]{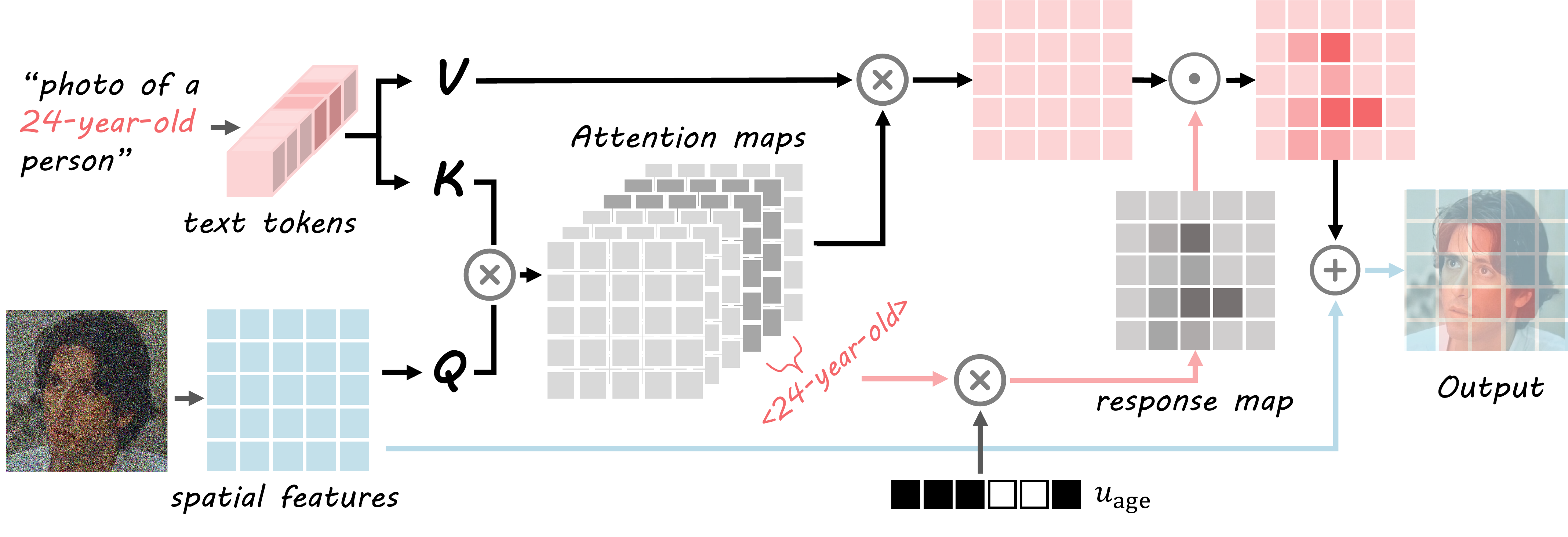}
  \caption{TTAB mechanism, using age-token attention maps as a
spatial response map to boost on age-relevant regions. }
  \label{fig4}
  \vspace{-0.5em}
\end{figure}
To address this, we propose TTAB mechanism to concentrate updates on regions associated with age semantics. As shown in Fig.~\ref{fig4}, within each UNet cross-attention block, an attention map $A\in \mathbb R^{(h\times w)\times n}$ is computed between linearly projected spatial features (as $Q$) and text tokens (as $K, V$), where $h\times w$ is the number of spatial locations, and n is the token length. Since $A[(i, j) ,k]$ quantifies how strongly the spatial location $(i,j)$ attends to the $k$-th text token, taking the slice $A[:,S_{age}]$, where $S_{age} \subseteq \{1,\ldots,n\}$ indexes the age-token subset (``<$\tau$-year-old>''), yields a spatial activation map highlighting age-sensitive regions. Building on this, we aggregate these attention weights to form a discriminative spatial weight map, defined as:
\begin{equation}
    u_{\text{age}} \in \{0,1\}^{n}, \qquad
    (u_{\text{age}})_k =
    \begin{cases}
        1, & k \in S_{\text{age}}, \\
        0, & \text{otherwise},
    \end{cases}
\end{equation}

\begin{equation}
   \gamma = \mathrm{norm}((QK^{\top})u_{\text{age}}),
\end{equation}
here, $u_{\text{age}}$ acts as a binary selector that selects age-related tokens and aggregates them into a single spatial response map $\gamma \in \mathbb{R}^{(h \times w) \times 1}$, which is then normalized to $[0,1]$ by the $\mathrm{norm}(\cdot)$ operation. This weight is broadcast along the channel dimension to modulate the attention output:
\begin{equation}
z = z + \gamma \odot \mathrm{softmax}\left( \frac{QK^{\top}}{\sqrt{d_c}} \right) V,
\end{equation}
where $z$ is the latent spatial feature. We operate on the $16^2$ cross-attention maps, which empirically contain the most semantic information  \cite{p2p, towards}. In this way, TTAB complements AAGG, effectively channeling the optimization flow into age-related areas without amplifying changes elsewhere. 

\section{Experiments}

\subsection{Experimental Setup}
\textbf{Training Datasets.} We train our model on two widely used identity-paired datasets, VGGFace2-HQ \cite{vggface2} and CelebRef-HQ \cite{DMDNet}. In total, 5,405 identities are selected, yielding 178,877 images. All images are centrally aligned and resized to a spatial resolution of $512^2$. For each training instance, we randomly sample 1$\sim$5 images with the same identity as the reference set. A commonly adopted first-order degradation pipeline is used to synthesize the corresponding low-quality inputs. Details are provided in the Appendix D.

\noindent
\textbf{Testing Datasets.} We evaluate our method under both same-age and cross-age settings: the same-age evaluation verifies the identity-preserving capability learned during training, following the standard protocol adopted in existing reference-based methods, while the cross-age evaluation examines our primary objective—achieving restorations that are faithful to both identity and age.
\begin{itemize}
    \item For \textbf{same-age} testing, we disable the plug-and-play age control generation (AAGG \& TTAB) and perform inference directly with the trained identity-preserving model. We select 150 identities from the remaining individuals in CelebRef-HQ \cite{DMDNet}, using the same setup as training and setting ``photo of a person'' as global prompt to focus evaluation on identity fidelity.
    \item For \textbf{cross-age} testing, we use one synthetic dataset and one real-world dataset.  The \textbf{synthetic dataset} is constructed from AgeDB \cite{agedb}, which is identity-divided and annotated with age labels for each image. The average age span per identity is 50.3 years. As the images are low-resolution and non-uniform, we first apply a super-resolution model \cite{sparnet} to upscale them to $512^2$, and further filter samples using ArcFace \cite{arcface} to ensure identity consistency. Following this process, we curate a subset of 100 identities, each with 1$\sim$5 reference images. Ground-truth age labels are used to construct age-specific prompts during inference. For the \textbf{real-world dataset}, we collect low-quality face images of 20 public figures from online sources and low-resolution video frames. The shooting age of each image is inferred from publicly available records, and for each identity, we gather high-quality images taken at least 20 years apart to serve as reference images. Low-quality inputs are generated using the same degradation pipeline as in training.
\end{itemize}

\noindent
\textbf{Implementation Details.} We use Stable Diffusion 2.1 \cite{sd} with the pretrained ControlNet from DiffBIR \cite{diffbir} as our backbone. The model is finetuned for 270K iterations using the AdamW \cite{adamw} optimizer with a learning rate of 4e-5. Training is conducted on 4 NVIDIA RTX 4090 GPUs with a batch size of 12 per GPU. 

\subsection{Comparisons with State-of-the-art Methods}
\begin{figure*}[t]
  \centering
  \includegraphics[width=0.999\linewidth]{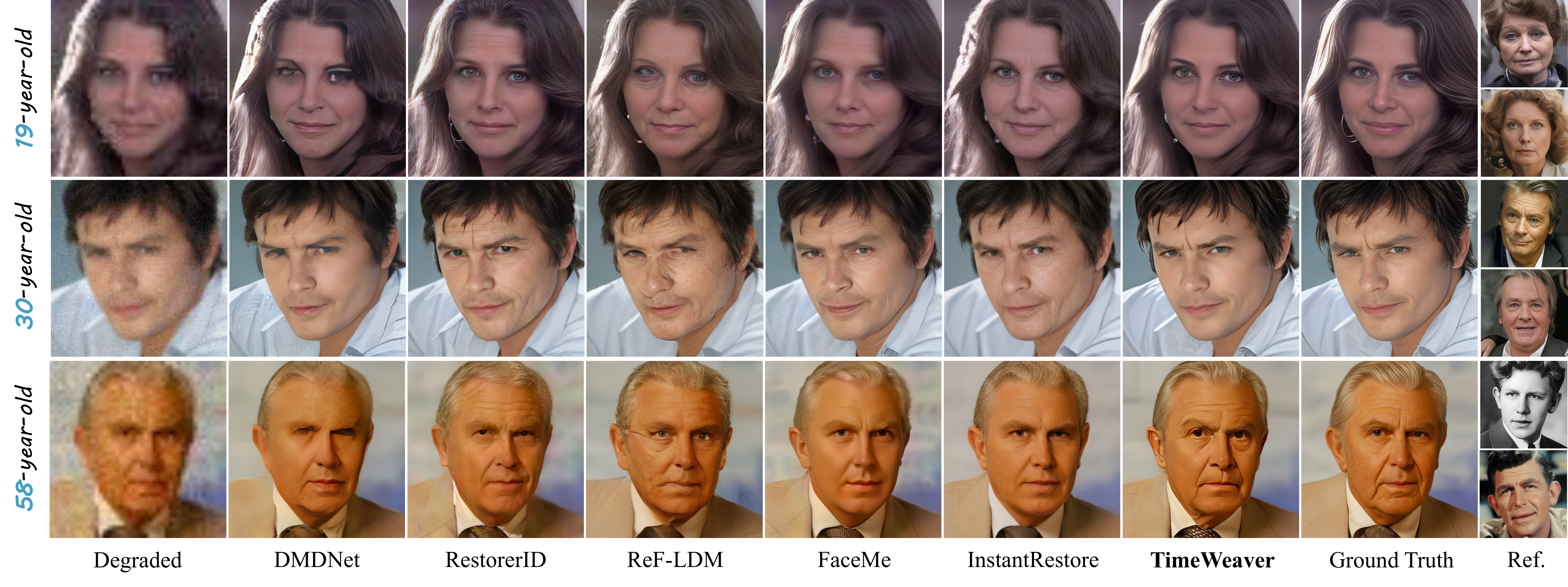} 
  \vspace{-1.6em}
  \caption{Qualitative comparison with reference-based methods on \textbf{synthetic dataset}. } 
  \label{fig5}
  \vspace{-0.2em}
\end{figure*}

\begin{figure*}[t]
  \centering
  \includegraphics[width=0.999\linewidth]{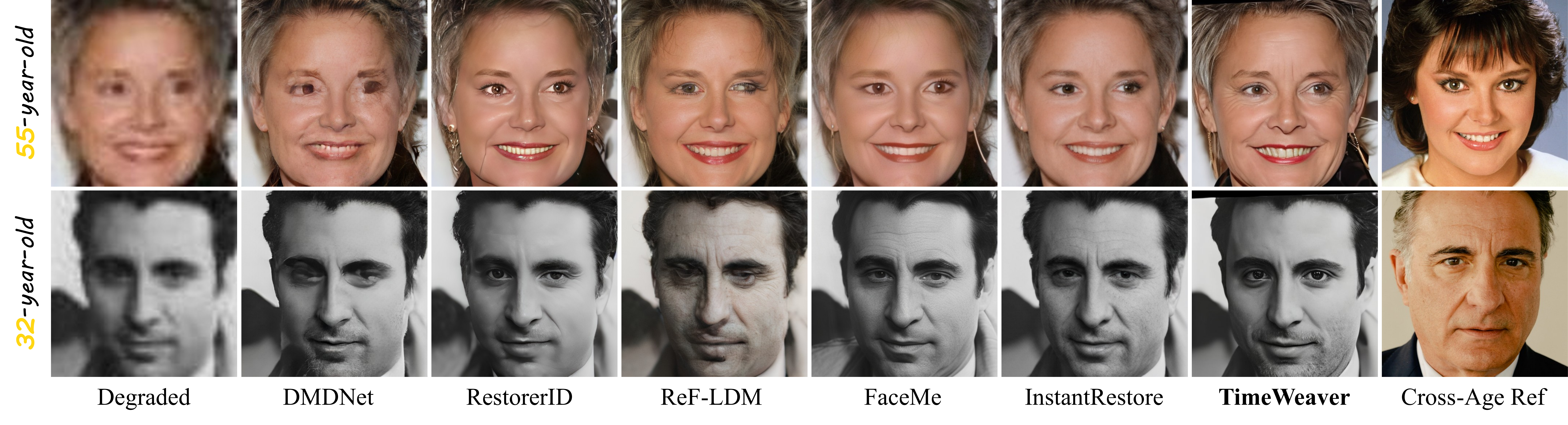} 
  \vspace{-1.6em}
  \caption{Qualitative comparison with reference-based methods on \textbf{real-world dataset}.} 
  \label{fig6}
  \vspace{-0.8em}
\end{figure*}
\textbf{Compared Methods.} We compare TimeWeaver with state-of-the-art baselines. For reference-free restoration, we include CodeFormer \cite{codeformer}, DifFace \cite{difface}, and DiffBIR \cite{diffbir}(base model). For reference-based methods, we consider all publicly available approaches with released code, namely DMDNet \cite{DMDNet}, RestorerID \cite{restorerid}, Ref-LDM \cite{ref-ldm}, FaceMe \cite{faceme}, and InstantRestore \cite{instantrestore}.

\noindent
\textbf{Evaluation Metrics.} The analysis is conducted from three perspectives: image quality, identity similarity, and age consistency. For synthetic datasets, image quality is assessed using PSNR, SSIM, and LPIPS \cite{lpips} (full-reference), as well as NIQE \cite{niqe} and FID \cite{fid} (no-reference), all computed via the PyIQA library. Identity similarity (IDS) is measured by the cosine similarity between ArcFace embeddings \cite{arcface}. Age consistency (AGE) is quantified by predicting the age of restored images using the DEX age estimator \cite{dex}—one of the most widely used and validated age-estimation baselines—and calculating the mean absolute error against ground-truth labels. For the real-world dataset, where ground-truth images are unavailable, we use NIQE, FID, IDS, and AGE metrics. In this case, IDS is computed between restored images and the reference images.

\noindent
\textbf{Evaluation on Same-age Dataset.} As shown in Tab.~\ref{tab1}(left), TimeWeaver achieves the best performance in LPIPS, NIQE, and IDS, showing superior perceptual quality, naturalness, and identity preservation. It also ranks second in SSIM and FID with scores close to the best method, while maintaining competitive PSNR. These results confirm that our model delivers strong general-purpose restoration quality comparable to, and often surpassing existing approaches in standard same-age scenarios. Qualitative comparisons are provided in Appendix E.

\noindent
\textbf{Evaluation on Cross-age Synthetic Dataset.} Tab.~\ref{tab1}(right) reports quantitative results on synthetic dataset AgeDB \cite{agedb}, where large age gaps exist between degraded inputs and reference images. Our method achieves top performance in NIQE, FID, IDS, and AGE, while maintaining competitive LPIPS scores, demonstrating superior identity fidelity and perceptual quality. Notably, it significantly surpasses all baselines in age accuracy, indicating superior age alignment capability. Fig.~\ref{fig5} clearly reveals that existing reference-based methods suffer from noticeable age drift, whereas ours effectively corrects age deviations. 

\noindent
\textbf{Evaluation on Cross-age Real-world Dataset.} For the real-world dataset, Tab.~\ref{tab2} reports quantitative comparisons. TimeWeaver achieves the best performance on NIQE and AGE, and ranks second in FID and IDS, with only a slight gap from the top method in IDS. We attribute this minor drop to subtle identity variations introduced by age changes, as IDS is computed against reference images. Fig.~\ref{fig6} further validates exceptional robustness under real-world degradations. TimeWeaver produces sharper structures, more faithful identity traits, and more natural age characteristics. In contrast, others exhibit issues such as color deviation, blurry artifacts, or noticeable age drift.

\begin{table*}[t]
\centering
\caption{Quantitative comparison on same-age and cross-age(synthetic) datasets. The best results are shown in \textbf{bold}, and the second-best are \underline{underlined}.}
\vspace{-0.6em}
\label{tab1}
\setlength{\tabcolsep}{4pt}
\resizebox{\textwidth}{!}{%
\begin{tabular}{l c | c c c c c | c | c c c c c | c | c}
\toprule

& & \multicolumn{6}{c|}{\textbf{Same-age}} & \multicolumn{7}{c}{\textbf{Cross-age (synthetic)}} \\[2pt]
\textbf{Method} & \textbf{Ref} &
PSNR $\uparrow$ & SSIM $\uparrow$ & LPIPS $\downarrow$ &
NIQE $\downarrow$ & FID $\downarrow$ & IDS $\uparrow$ &
PSNR $\uparrow$ & SSIM $\uparrow$ & LPIPS $\downarrow$ &
NIQE $\downarrow$ & FID $\downarrow$ & IDS $\uparrow$ & AGE $\downarrow$ \\
\midrule

CodeFormer \cite{codeformer}    
&   & 25.61 & 0.720 & 0.207 & 4.56 & 47.46 & 0.639 
& \textbf{26.44} & \textbf{0.756} & 0.207 & 4.60 & 57.46 & 0.639 & 11.39 \\

DifFace \cite{difface}       
&   & 25.31 & 0.720 & 0.247 & \textbf{4.20} & 55.01 & 0.509 
& 24.95 & 0.694 & 0.268 & \underline{4.21} & 75.12 & 0.432 & \underline{10.66} \\

DiffBIR \cite{diffbir}       
&   & \underline{26.03} & 0.705 & 0.220 & 6.33 & 56.55 & 0.698 
& 25.79 & 0.679 & 0.220 & 6.22 & 52.99 & 0.512 & 12.25 \\

DMDNet \cite{DMDNet}        
& \checkmark & 25.76 & 0.730 & 0.228 & \underline{4.45} & 55.48 & 0.703 
& 25.85 & 0.717 & 0.224 & 4.40 & 56.74 & 0.645 & 12.75 \\

RestorerID \cite{restorerid}    
& \checkmark & 25.00 & 0.699 & 0.272 & 4.92 & 59.74 & 0.686 
& 25.12 & 0.687 & 0.231 & 4.66 & 59.76 & 0.644 & 15.00 \\

Ref-LDM \cite{ref-ldm}       
& \checkmark & 24.80 & 0.713 & 0.227 & 4.69 & 46.46 & 0.714 
& 24.73 & 0.684 & 0.225 & 4.64 & 55.75 & 0.625 & 18.49 \\

FaceMe \cite{faceme}        
& \checkmark & \textbf{26.41} & 0.733 & 0.220 & 4.92 & 47.34 & 0.722 
& \underline{26.37} & 0.723 & 0.220 & 5.31 & \underline{51.97} & 0.648 & 22.52 \\

InstantRestore \cite{instantrestore} 
& \checkmark & 25.60 & \textbf{0.747} & \underline{0.199} & 6.06 & \textbf{46.06} & \underline{0.725} 
& 25.77 & \underline{0.731} & \textbf{0.199} & 6.42 & 53.52 & \underline{0.692} & 16.61 \\

\rowcolor{RowColor}
\textbf{TimeWeaver} (ours) 
& \checkmark 
& 25.50 & \underline{0.735} & \textbf{0.198} & \textbf{4.20} & \underline{46.30} & \textbf{0.738} 
& 25.11 & 0.718 & \underline{0.201} & \textbf{3.52} & \textbf{51.28} & \textbf{0.701} & \textbf{8.25} \\

\bottomrule
\end{tabular}}
\vspace{-1em}
\end{table*}

\noindent
\textbf{User Study.} Face restoration is inherently human-centric, especially when evaluating subtle attributes such as identity and age. As objective metrics cannot fully reflect human perception, we conduct a user study for a more reliable assessment. We compare our method against five competitive baselines: CodeFormer \cite{codeformer}, RestorerID \cite{restorerid}, Ref-LDM \cite{ref-ldm}, FaceMe \cite{faceme}, and InstantRestore \cite{instantrestore}. A total of 50 volunteers participate in the study. The questionnaire covered three dimensions: visual quality, identity similarity, and age consistency (provided in Appendix H). The results are summarized in Fig.~\ref{fig7}. Although TimeWeaver scores slightly lower than CodeFormer in visual quality (26.0\% vs. 26.5\%), it substantially outperforms all baselines in identity similarity (37.8\%) and achieves a significant lead in age consistency, receiving 64.5\% of all votes, which is +45.6 percentage points higher than the second-best method. These results indicate that TimeWeaver maintains high visual quality while delivering the most faithful identity preservation and age control in cross-age restoration.

\begin{table}[t]
\centering
\captionof{table}{Quantitative comparison on the real-world dataset. }
\label{tab2}
\resizebox{0.99\linewidth}{!}{
\setlength{\tabcolsep}{6pt}
\renewcommand{\arraystretch}{1.1}
\begin{tabular}{l c|c c|c|c}
\toprule
\textbf{Method} & \textbf{Ref} & NIQE $\downarrow$ & FID $\downarrow$ & IDS $\uparrow$ & AGE $\downarrow$ \\
\midrule
CodeFormer \cite{codeformer}  &   & 4.19 & \textbf{56.53} & 0.263 & 9.67 \\
DifFace \cite{difface}      &      & \underline{3.87} & 119.78 & 0.295 & 11.83 \\
DiffBIR \cite{diffbir}     &       & 6.07 & 67.97 & 0.290 & 10.50 \\
DMDNet \cite{DMDNet}     & \checkmark        & 4.35 & 105.54 & 0.295 & 9.67  \\
RestorerID \cite{restorerid} & \checkmark    & 4.40 & 76.31  & 0.467 & 15.14 \\
Ref-LDM \cite{ref-ldm}     & \checkmark      & 4.31 & 59.93  & 0.352 & 13.33 \\
FaceMe \cite{faceme}      & \checkmark       & 4.80 & 80.99  & 0.372 & \underline{9.00}  \\
InstantRestore \cite{instantrestore} & \checkmark  & 5.84 & 99.79  & \textbf{0.515} & 16.67 \\
\rowcolor{RowColor}\textbf{TimeWeaver} (ours)  & \checkmark
& \textbf{3.84} & \underline{58.21} & \underline{0.507} & \textbf{7.16} \\
\bottomrule
\end{tabular}
} 
\end{table}

\begin{figure}[t]
    \includegraphics[width=0.99\linewidth]{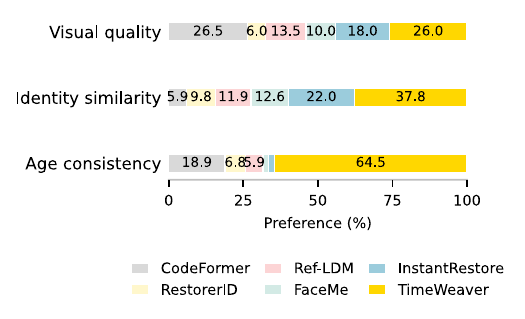}
\caption{User study results.}
\label{fig7}
\end{figure}

\subsection{Comparisons with Two-stage Restoration-Editing Pipelines}
To examine whether cross-age restoration can be achieved through post-hoc editing, we take the restored outputs of four reference-based restoration methods—RestorerID \cite{restorerid}, Ref-LDM \cite{ref-ldm}, FaceMe \cite{faceme}, and InstantRestore \cite{instantrestore}—and apply age-editing methods as a second-stage adjustment using age-specific prompt as the target editing condition. Specifically, we use three dedicated age-editing models (HRFAE \cite{frfae}, SAM \cite{sam}, FADING \cite{fading}) and three general-purpose diffusion-based image editing methods (Null-text Inversion + Prompt2prompt \cite{null}, DDS \cite{dds}, FLUX.1 Kontext \cite{flux}), where DDS is similar to our inference strategy. 

\begin{figure*}[t]
  \centering
  \includegraphics[width=0.999\linewidth]{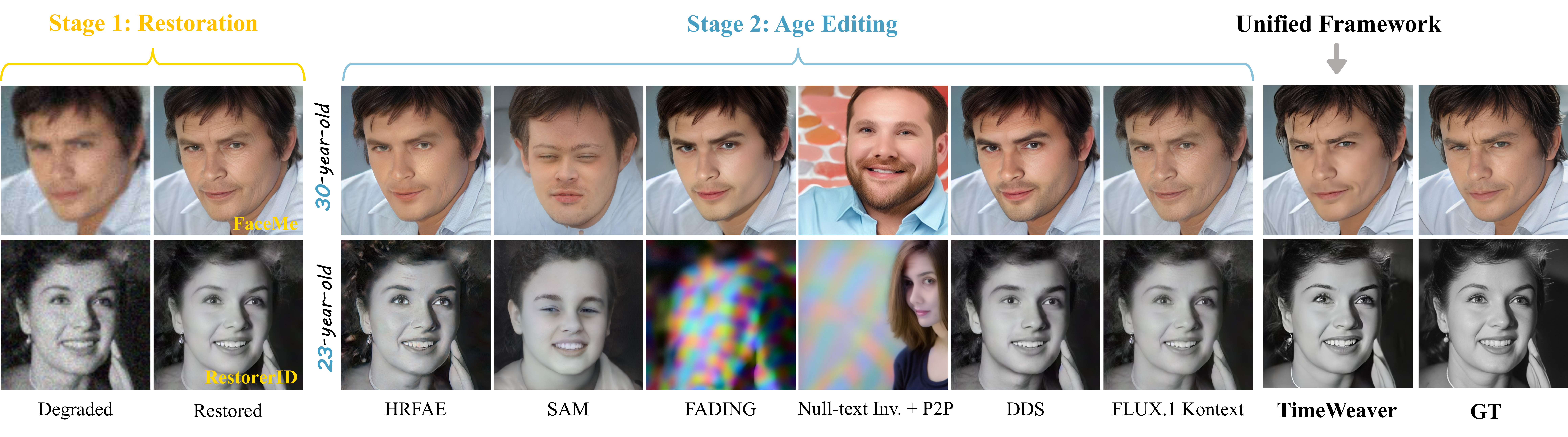} 
  \vspace{-1.6em}
  \caption{Qualitative comparison with Two-stage Restoration–Editing Pipelines} 
  \label{fig8}
\end{figure*}

As shown in Fig.~\ref{fig8}, the editing results reveal clear limitations of two-stage pipeline. Age-editing methods often distort facial structures (SAM \cite{sam}) or introduce artifacts (FADING \cite{fading}). The inversion-based method \cite{null} can drastically alter the image content. General-purpose editing (DDS \cite{dds}, FLUX.1 Kontext \cite{flux}) struggle with fine-grained attributes like age, leading to incomplete age changes and sometimes causing semantic misunderstandings (\emph{e.g.}, gender changes). 

Tab.~\ref{tab3} presents quantitative results of applying editing methods to the outputs of FaceMe \cite{faceme}, further supporting these observations: all editing methods exhibit significant degradation in visual quality and identity similarity when applied to restored images and fail to correct the target age. In contrast, our unified framework avoids the error accumulation of two-stage pipelines and enables reliable cross-age restoration while maintaining visual quality and identity fidelity.

\begin{table}[t]
\centering
\caption{Comparison with editing methods. }
\label{tab3}
\vspace{-0.3em}
\setlength{\tabcolsep}{3pt}
\renewcommand{\arraystretch}{1.24}
\resizebox{0.98\linewidth}{!}{
\begin{tabular}{l | c | c c c | c | c | c}
\toprule
\textbf{Method} & Type & PSNR $\uparrow$ & LPIPS $\downarrow$ & NIQE $\downarrow$ & IDS $\uparrow$ & AGE $\downarrow$ & Time(s) $\downarrow$ \\
\midrule
HRFAE \cite{frfae} & GAN & 21.57 & 0.308 & 5.31 & 0.602 & 10.23 & \textbf{0.18} \\
SAM \cite{sam} & GAN & 18.09 & 0.457 & 12.33 & 0.347 & 8.77 & \underline{0.45} \\
FADING \cite{fading} & Diffusion & 22.28 & 0.270 & 6.52 & 0.584 & \underline{8.05} & 133.41 \\
Null Inv. + P2P \cite{null} & Diffusion & 20.91 & 0.225 & 8.90 & 0.401 & 14.54 & 57.26 \\
DDS \cite{dds} & Diffusion & 23.60 & \underline{0.207} & 5.08 & 0.623 & 15.67 & 12.70 \\
FLUX.1 Kontext \cite{flux} & Diffusion & \underline{24.37} & 0.208 & \underline{4.67} & \underline{0.643} & 13.40 & 40.12 \\
\rowcolor{RowColor}
TimeWeaver (ours) & Diffusion & \textbf{25.11} & \textbf{0.201} & \textbf{3.52} & \textbf{0.701} & \textbf{8.25} & 22.68 \\
\bottomrule
\end{tabular}}
\end{table}

\begin{table}
\centering
\caption{Performance across different age gaps.}
\label{tab4}
\vspace{-0.3em}
\setlength{\tabcolsep}{2pt}
\renewcommand{\arraystretch}{1.2}
\resizebox{\linewidth}{!}{
\begin{tabular}{l | c c c c c|c|c}
\toprule
\textbf{Age Gap} &
PSNR $\uparrow$ &
SSIM $\uparrow$ &
LPIPS $\downarrow$ &
NIQE $\downarrow$ &
FID $\downarrow$ &
IDS $\uparrow$ &
AGE $\downarrow$ \\
\midrule
$\le 10$ & 24.58 & 0.698 & 0.203 & 3.81 & 53.08 & 0.683 & 5.66 \\
10 $\sim$ 20 & 25.27 & 0.708 & 0.203 & 3.92 & 53.36 & 0.687 & 5.41 \\
20 $\sim$ 30 & 25.12 & 0.716 & 0.212 & 3.80 & 54.43 & 0.682 & 5.89 \\
30 $\sim$ 40 & 24.50 & 0.692 & 0.201 & 3.86 & 55.67 & 0.677 & 6.33 \\
$> 40$ & 24.23 & 0.700 & 0.217 & 4.00 & 55.98 & 0.670 & 6.70 \\
Mixed & 24.37 & 0.712 & 0.208 & 3.86& 54.60 & 0.682 & 6.55 \\
\bottomrule
\end{tabular}}

\end{table}

\subsection{Ablation Studies}
\textbf{Robustness on Varying Age Gaps.} To assess the robustness of our method under different age discrepancy between the degraded and its references, we select 30 identities from the cross-age dataset. For each identity, reference images are divided into five age-gap intervals relative to the degraded input, yielding five restorations with increasing age disparities, plus an additional mixed-age setting where references come from multiple age stages. As shown in Tab.~\ref{tab4}, all metrics remain stable across intervals, demonstrating that TimeWeaver is resilient to large age gaps without compromising visual quality, identity, or age consistency.

\begin{table}[t]
\centering
\caption{Ablation study on different identity representations.}
\label{tab5}
\vspace{0.4em}
\setlength{\tabcolsep}{2pt}
\renewcommand{\arraystretch}{1.5}
\resizebox{\linewidth}{!}{
\begin{tabular}{l | c c c c c|c|c}
\toprule
\textbf{Method} &
PSNR $\uparrow$ &
SSIM $\uparrow$ &
LPIPS $\downarrow$ &
NIQE $\downarrow$ &
FID $\downarrow$ &
IDS $\uparrow$ &
AGE $\downarrow$ \\
\midrule
w/o $f_{global}$ & 24.47 & 0.645 & 0.247 & 5.60 & 60.00 & 0.568 & 11.30 \\
w/o $f_{facial}$ & 25.01 & 0.689 & 0.223 & 4.45 & 58.98 & 0.687 & 8.79 \\
w/ $f_{global}$ \& $f_{facial}$ (no mask) & \textbf{25.63} & 0.713 & 0.218 & \textbf{3.40} & 54.76 & \textbf{0.711} & 10.43 \\
\rowcolor{RowColor} w/ $f_{global}$ \& $f_{facial}$ (mask)
& 25.11 & \textbf{0.718} & \textbf{0.201} & 3.52 & \textbf{51.28} & 0.701 & \textbf{8.25} \\
\bottomrule
\end{tabular}}
\end{table}

\begin{table}
\caption{Ablation study on AAGG and TTAB mechanisms.}
\label{tab6}
\vspace{0.4em}

\setlength{\tabcolsep}{2pt}
\renewcommand{\arraystretch}{1.5}
\resizebox{\linewidth}{!}{
\begin{tabular}{l | c c c c c|c|c}
\toprule
\textbf{Method} & PSNR $\uparrow$ & SSIM $\uparrow$ & LPIPS $\downarrow$ &
NIQE $\downarrow$ & FID $\downarrow$ & IDS $\uparrow$ & AGE $\downarrow$ \\
\midrule
w/o AAGG  & 25.08 & 0.711 & 0.208 & 4.56 & 52.97 & 0.677 & 14.43 \\
w/o TTAB  & 23.34 & 0.670 & 0.212 & 3.80 & 56.67 & 0.698 & 9.06 \\
\rowcolor{RowColor} w/ AAGG \& TTAB
& \textbf{25.11} & \textbf{0.718} & \textbf{0.201} & \textbf{3.52} & \textbf{51.28} & \textbf{0.701} & \textbf{8.25} \\
\bottomrule
\end{tabular}}
\end{table}

\noindent
\textbf{Effect of Identity Representation.} We investigate the effect of different identity representations during training. Specifically, we ablate the global identity embedding $f_{global}$, the facial tokens $f_{facial}$, and the patch-level masking strategy. As shown in Tab.~\ref{tab5}, removing $f_{global}$ (row 1) causes a clear drop in IDS, indicating its key role in identity preservation. Introducing $f_{facial}$ (row 2--4) improves visual quality metrics (PSNR, SSIM, LPIPS, NIQE, and FID), confirming that $f_{facial}$ complements $f_{global}$ with fine-grained details. The last two rows reveals a trade-off between identity similarity (IDS) and age consistency (AGE). Without masking, the model achieves the highest IDS, but AGE increases, indicating that age-related cues leak into $f_{facial}$. Masking suppresses this leakage and yields a lower AGE while maintaining strong IDS (0.701). 

\begin{figure}[t]
    \centering
    \begin{subfigure}{0.98\linewidth}
        \centering
        \includegraphics[width=\linewidth]{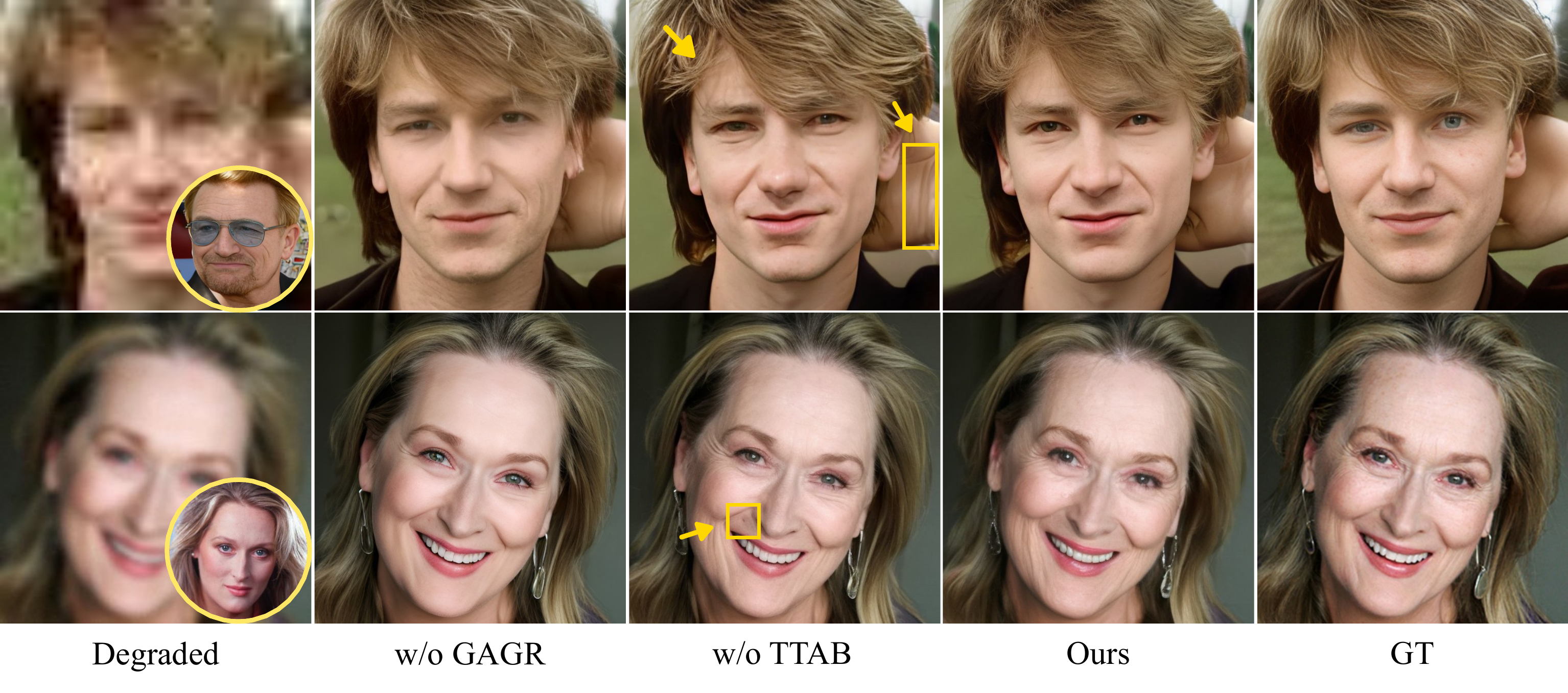}
        \caption{Visual comparison of AAGG and TTAB.}
        \vspace{0.2em}
        \label{fig9a}
    \end{subfigure}
    \begin{subfigure}{0.88\linewidth}
        \centering
        \includegraphics[width=\linewidth]{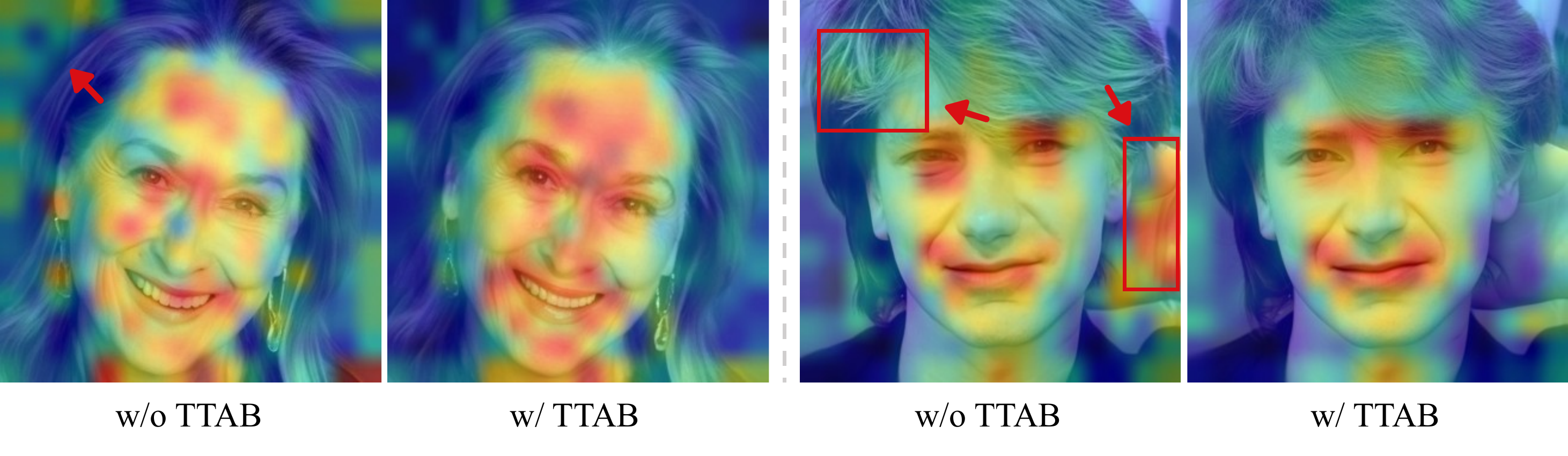}
        \caption{Visualization of cross-attention maps for the age token.}
        \vspace{0.2em}
        \label{fig9b}
    \end{subfigure}
    \vspace{-0.6em}
    \caption{Visual comparisons of ablation studies.}
    \label{fig9}
    \vspace{-1.5em}
\end{figure}

\noindent
\textbf{Effect of Age-Aware Gradient Guidance.} Tab.~\ref{tab6} shows the impact of AAGG at inference. Removing it leads to a noticeable increase in AGE (8.25 → 14.43), indicating that AAGG is the key driver of accurate age alignment. Fig.~\ref{fig9}(a) further validates this observation. Although identity conditioning mitigates reference age information, it cannot fully remove it; without AAGG, residual age cues may still be utilized. AAGG refines the process by enforcing age control, enabling precise age-specific editing.

\noindent
\textbf{Effect of Token-Targeted Attention Boost.} From Tab.~\ref{tab6} (rows 2 and 3), removing TTAB degrades image quality, reflected by noticeable worse PSNR, SSIM, and FID. Qualitatively (Fig.~\ref{fig9}(a)), its absence introduces texture fluctuations such as over-sharpened hair/skin and spurious mole that is not faithful to the input. We interpret this to the global updates applied by AAGG, which tend to spread across the entire image. TTAB redirects updates toward age-relevant regions, thereby alleviating off-target perturbations. Consistently, Fig.~\ref{fig9}(b) visualizes the attention map of the age token (“$\tau$-year-old”), showing more concentrated activation on age-related regions with TTAB, confirming that it enables localized age editing instead of globally altering textures.
\section{Conclusion}
We present TimeWeaver, the first reference-based face restoration framework capable of handling cross-age references. By decoupling identity learning and age control, TimeWeaver presents an age-robust identity representation and employs AAGG and TTAB at inference for precise age editing. Extensive experiments demonstrate state-of-the-art performance in visual quality, identity fidelity, and age consistency.

{
    \small
    \bibliographystyle{ieeenat_fullname}
    \bibliography{main}
}


\end{document}